\documentclass[letterpaper,10pt,journal]{IEEEtran}
\usepackage[margin=0.6in]{geometry}
\usepackage{graphicx,amsmath,amssymb,booktabs,array,tabularx,times,url,siunitx}
\usepackage[bookmarks=false,hidelinks]{hyperref}
\let\url\nolinkurl
\newcommand{\email}[1]{\url{#1}}
\hypersetup{pdftitle={Outcome-Conditioned End-Effector Geometry Across Vision-Language-Action Policies},pdfauthor={Xingyu Lin and Zhuang Li and Zhongrun Wu and Shouquan Zhou and Dehui Du}}
\graphicspath{{figures/}}
\title{\LARGE\bf Outcome-Conditioned End-Effector Geometry\\Across Vision-Language-Action Policies}
\author{Xingyu Lin\textsuperscript{1}, Zhuang Li\textsuperscript{2}, Zhongrun Wu\textsuperscript{3}, Shouquan Zhou\textsuperscript{4}, and Dehui Du\textsuperscript{5}%
\thanks{\textsuperscript{1}School of Software Engineering, East China Normal University, Shanghai, China (e-mail: \email{xingyulin@stu.ecnu.edu.cn}).}%
\thanks{\textsuperscript{2}School of Software Engineering, East China Normal University, Shanghai, China (e-mail: \email{dfei30419@gmail.com}).}%
\thanks{\textsuperscript{3}School of Computer Science and Technology, East China Normal University, Shanghai, China (e-mail: \email{r2668940489@gmail.com}).}%
\thanks{\textsuperscript{4}School of Maritime Economics and Management, Dalian Maritime University, Dalian, China (e-mail: \email{zhoushouquan020628@gmail.com}).}%
\thanks{\textsuperscript{5}School of Software Engineering, East China Normal University, Shanghai, China (e-mail: \email{dhdu@sei.ecnu.edu.cn}).}}
\begin{document}
\bstctlcite{paper:bstcontrol}
\maketitle
\thispagestyle{empty}\pagestyle{empty}
\begin{abstract}
Vision-language-action (VLA) policies solve the same manipulation task through different action interfaces, but task success alone does not establish whether their physical executions agree. We study cross-policy end-effector geometry in 15,000 closed-loop LIBERO rollouts from four policies. The primary clean-condition analysis forms 3,600 configuration-matched, and therefore dependent, policy pairs. Both-success pairs have a median normalized dynamic time warping distance of 0.0120\,m versus 0.0380\,m when exactly one policy succeeds. This ordering holds in every task, every policy pair, and nine sampling and band-limited representations; however, the ratio varies severalfold across representations, so we report the direction rather than a fixed multiple. Both-failure pairs are more separated again but rest on thin, uneven support, so we report them as exploratory. Within successful executions, partner replacements separate more across tasks than across initial states. A matched baseline still reveals measurable, heterogeneous residual policy differences, so a low cross-policy distance does not imply interchangeability. Successful executions sit about as far from same-task demonstrations as those demonstrations sit from each other, compatible with task-associated geometry without separating training-data overlap from task constraints. A common 72-action window preserves the ordering but reduces its magnitude; endpoint and duration adjustment likewise leaves a positive mixed-outcome coefficient relative to both-success pairs, though its magnitude is specification-dependent. Under composite visual stress, policy rankings and pair composition change together.
\end{abstract}

\begin{figure*}[t]
\centering\includegraphics[width=\textwidth]{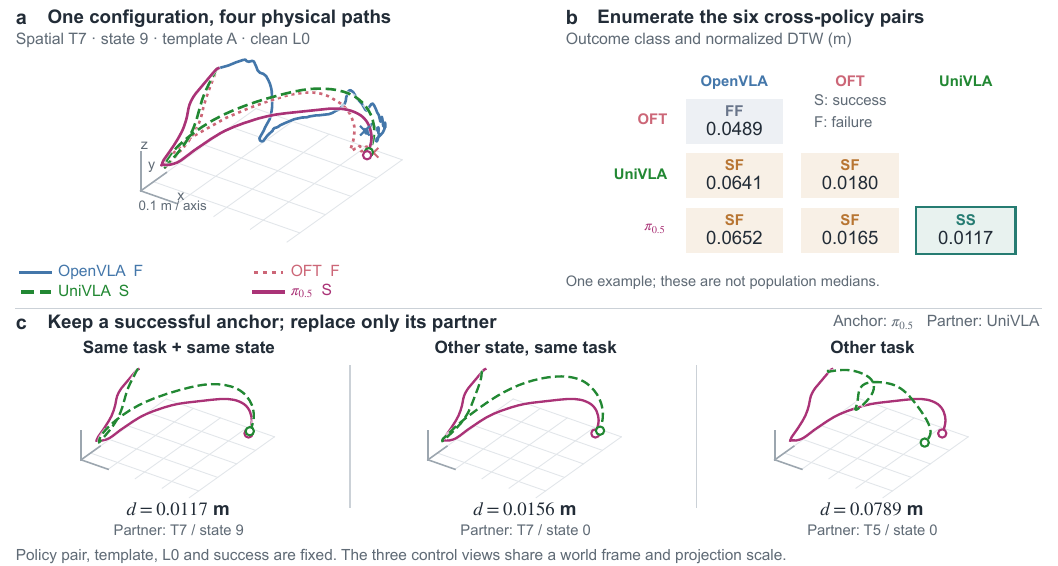}
\caption{Primary comparison design. (a) Recorded end-effector paths from a selected clean configuration with two successes and two failures. (b) The six unordered cross-policy pairs; blank cells omit self- and duplicate pairings, and distances describe this example, not population medians. (c) A successful $\pi_{0.5}$ anchor is retained while its UniVLA partner changes state or task, holding policy pair, template, level, and successful outcomes fixed. Views are orthographic projections; DTW uses the original 3D coordinates.}
\label{fig:overview}
\end{figure*}

\section{Introduction}
Vision-language-action (VLA) policies map images and task instructions to robot actions, and their action interfaces now diverge sharply. OpenVLA produces discretized action tokens~\cite{openvla}, OpenVLA-OFT (OFT) predicts continuous action chunks~\cite{openvla_oft}, UniVLA decodes learned latent actions~\cite{univla}, and $\pi_{0.5}$ uses a flow-based action model~\cite{pi05}. These policies are compared almost exclusively by a task success rate. A leaderboard number, however, is silent about how the arm actually moves: two policies that each succeed on 90\% of trials can reach the goal through different routes, and two that fail can diverge or fail after following similar routes.

This gap matters in deployment: replacing one policy with another at the same success rate can silently change motion and workspace use. It also matters scientifically, because success counts do not show whether heterogeneous VLA pipelines trace similar end-effector paths toward a goal or follow distinct paths. Recent work enriches evaluation with per-policy behavioral metrics and robustness profiles~\cite{roboeval,nebula,uncertaintyquality,distractedrobot,liberoplus}, comparing policies mainly through per-policy scores or the sets of scenes each solves. We ask a complementary question about \emph{pairs}: under a matched task, initial state, instruction, and visual condition, how close are the end-effector paths of two different policies, and how does that closeness depend on their joint outcome? Task success alone does not specify how closely two end-effector paths agree. We measure their agreement and its sensitivity to endpoint proximity and episode duration, in an offline comparison rather than online failure prediction or safety certification.

Conditioning on the joint outcome is central, not cosmetic. Each pair enters a both-success (SS), mixed (SF), or both-failure (FF) stratum. Pooling by ``agreeing outcome'' is misleading because successes dominate that pool, so a low pooled distance would describe successful execution while appearing to describe failure. Comparing successes against failures also changes endpoints, termination times, and task completion---properties of the measurement problem, not nuisances that alignment removes.

On 3{,}600 configuration-matched pairs from the clean LIBERO-Spatial grid~\cite{libero}, SS paths have a median distance of 0.0120\,m versus 0.0380\,m for SF. This ordering holds in every task, policy pair, and representation examined, but a common observation window attenuates the gap. Within successful executions, matched controls locate the agreement as largely task-associated, while a within-policy baseline reveals small, heterogeneous residual policy differences rather than equivalence. FF has a larger median (0.0439\,m), but its thin support and representation-dependent rank relative to SF make it exploratory.

Our contribution is this finding and the comparison design that supports it: outcome-conditioned strata; matched task/state controls; a within-policy baseline; a demonstration reference at matched task identity; task--state block resampling; fixed observation windows; explicit alignment sensitivity; and an analysis of how visual stress changes conditional geometry and which pairs enter the comparison.

\section{Related Work}
\subsection{Behavioral Evaluation and Robustness}
VLATest mutates manipulation scenes and instructions~\cite{vlatest}, LIBERO-Plus expands evaluation across environmental and input dimensions~\cite{liberoplus}, NEBULA separates capability tests from operational probes such as timing and action stability~\cite{nebula}, RoboEval adds per-policy behavioral metrics that discriminate policies of similar success rate~\cite{roboeval}, and Distracted Robot reports distinct vulnerabilities and low agreement on which cluttered scenes policies solve~\cite{distractedrobot}. Each characterizes one policy, or its set of solved scenes, at a time. We instead compare the paired end-effector paths of two different policies at a shared configuration and condition every comparison on their joint outcome; our composite image ladder serves only to expose how performance and pair composition change within one recorded grid.

Benchmark audits of narrow evaluation distributions~\cite{benchmarkaudit} and deployment reports of differing behavior across implementations~\cite{experiences} motivate retaining task-level outcomes and reporting the execution interface rather than treating a policy name as a complete experimental specification.

Language-focused benchmarks provide stronger controls than a few fixed templates. LIBERO-Para decomposes meaning-preserving linguistic variation over object and action axes~\cite{liberopara}, multilingual evaluation examines the changing influence of language during execution~\cite{multilingual}, LIBERO-CF changes feasible instructions under familiar layouts to probe visual dominance~\cite{liberocf}, and wording-sensitivity studies show action outputs shift with instruction phrasing~\cite{wording}. Our template contrast supplies limited supporting evidence about one checkpoint; the primary controls ask how recorded physical paths differ when task or state matching changes.

\subsection{Physical Trajectories and Failure Signals}
Valle et al. evaluate action uncertainty and physical execution quality of individual runs~\cite{uncertaintyquality}; Kr\"uger et al. compare VLA inspection trajectories against reference paths in position and orientation~\cite{featurefollowing}; and dynamic time warping (DTW) itself is an established sequence-alignment method~\cite{dtw}. Our comparison object is two executed end-effector paths from different policies, grouped by joint outcome and compared against successful task/state replacements.

The closest positional comparison appears in LIBERO-Para's trajectory appendix, which builds a pseudo-reference from successful runs, resamples paths, applies DTW to end-effector positions, compares by outcome, and thresholds the distance to label failure types~\cite{liberopara}. We therefore claim no novelty in positional DTW comparison, resampling, or outcome-based grouping. Four differences define the increment. \emph{Object}: matched executions of two different policies, not one execution against a successful pseudo-reference. \emph{Conditioning}: the joint outcome of both executions, not one execution's outcome. \emph{Controls}: success-preserving task and state replacements, plus a within-policy baseline. \emph{Inference}: descriptive conditional geometry, with no distance threshold classifying a failure mechanism.

Several methods derive failure signals from policy representations, uncertainty, or information-theoretic quantities~\cite{safe,hideseek,triinfo}; these internal signals and our physical-path distances describe different spaces, since latent agreement can coexist with path separation. Our fixed-window analysis tests whether a geometric contrast is observable within a fixed horizon; it evaluates no detector.

\section{Study Design and Measurements}
\subsection{Tasks, Policies, and the Archive}
The common benchmark is LIBERO-Spatial: ten tasks, the first twenty states in each task's initial-state bank, three instruction templates, and five visual levels. Each policy contributes 3,000 rollouts, for 12,000 Spatial executions; a further 3,000 Object rollouts cover $\pi_{0.5}$ only, so cross-policy geometry is restricted to Spatial. The data have three units: an episode is one policy execution, a configuration fixes task, initial state, template, and visual level, and a comparison is one unordered pair of policies at that configuration. At L0, 600 configurations yield 2,400 episodes and 3,600 pairs; over L0--L4 the counts are 3,000, 12,000, and 18,000. Every policy/configuration cell has exactly one archived execution (Fig.~\ref{fig:overview}).

Spatial instructions use a task-specific relation $r$ in three templates (e.g., A, ``pick up the black bowl $r$ and place it on the plate''); Object templates analogously name an object and the basket.

The collector renders $256\times256$ observations, takes ten initialization steps, and permits up to 240 control actions, recording end-effector positions, emitted actions, gripper state, and tracked-object positions after each step. Success is the simulator's task-completion signal and terminates the episode; all archived failures reach the control cap. Array lengths define time, and initialization records hold zero action vectors. The archive contains 15,000 distinct keys and matching trajectory files; we checked array dimensions, finite values, record lengths, and source-file identity.

\begin{table}[t]
\centering
\caption{Interfaces used by the collector. E/W denote external/wrist views and P denotes proprioception. The pixel grid is where visual corruption is applied. Cadence is executed actions per policy query.}
\label{tab:interfaces}
\setlength{\tabcolsep}{3pt}
\begin{tabular*}{\columnwidth}{@{\extracolsep{\fill}}llcc@{}}
\toprule Policy & Inputs & Pixel grid & Cadence \\
\midrule
OpenVLA & E & $224^2$ & 1 \\
OFT & E,W,P & $256^2$ & 8 \\
UniVLA & E & $224^2$ & 1 \\
$\pi_{0.5}$ & E,W,P & $256^2$ & 5 \\
\bottomrule
\end{tabular*}
\end{table}

The first three policies use Spatial-finetuned checkpoints; the $\pi_{0.5}$ launcher selects its LIBERO configuration. OpenVLA and UniVLA query every control step. OpenVLA decodes tokens deterministically; UniVLA retains latent-action history and samples at temperature 0.75, top-$p$ 0.9. OFT uses a deterministic L1-trained continuous head with an eight-action queue; $\pi_{0.5}$ samples a flow-based chunk and executes five actions before replanning (Table~\ref{tab:interfaces}). OpenVLA and OFT therefore decode without sampling, UniVLA and $\pi_{0.5}$ are stochastic. Training data, checkpoints, horizons, and preprocessing also differ from one another and from official recipes, so these results describe the recorded implementations.

\subsection{Composite Visual Operators}
L0 leaves observations unchanged. Higher levels combine brightness scaling, Gaussian pixel noise, and wraparound shifts that alter content without moving the camera. From L2 onward, they also add possibly overlapping square masks---grayscale 128 at L2 and random at L3--L4---before clipping pixels to [0,255] (Table~\ref{tab:corruption}). These are fixed nominal settings, not an adaptive search. OpenVLA and UniVLA receive corruption after resizing while OFT and $\pi_{0.5}$ receive it on the original render, so equal mask dimensions cover different area fractions across policies; both queried views are corrupted for two-view policies.

\begin{table}[t]
\centering
\caption{Final operators per visual level. Brightness is uniform on the interval. Noise is in 8-bit pixel units; shift is in pixels. Mask notation gives count and side length.}
\label{tab:corruption}
\setlength{\tabcolsep}{2pt}
\begin{tabular*}{\columnwidth}{@{\extracolsep{\fill}}lcS[table-format=2.0]cc@{}}
\toprule Level & Brightness & {$\sigma$} & Shift & Masks \\
\midrule
L0 & 1 & 0 & 0 & none \\
L1 & [0.8,1.2] & 10 & $\pm8$ horiz. & none \\
L2 & [0.6,1.4] & 20 & $\pm16$ both & $1\times40^2$ \\
L3 & [0.45,1.6] & 35 & $\pm24$ both & $2\times60^2$ \\
L4 & [0.3,1.9] & 55 & $\pm32$ both & $3\times80^2$ \\
\bottomrule
\end{tabular*}
\end{table}

Each rollout's corruption seed hashes suite, policy, task, template, level, and state. Because we saved neither realized seeds nor queried images, matching a nonzero level matches operator settings, not the realized corruption across policies. The stress comparison therefore concerns these complete pipelines, folding together decoder, input, preprocessing, and cadence.

\subsection{Outcome-Conditioned Geometry}
Let $X_{m,c}$ be the recorded end-effector path of policy $m$ at configuration $c$, and let $q_{m,c}\in\{0,1\}$ denote its observed outcome. For every configuration, we enumerate all six unordered policy pairs. A pair is SS when $q_{m,c}+q_{n,c}=2$, SF when the sum is 1, and FF when it is 0, so the strata are defined by executions rather than by permanent groups of policies (Fig.~\ref{fig:overview}(b)).

For each path, retain every fifth recorded position in the common simulator frame. Given sampled sequences $X=(x_1,\ldots,x_{L_X})$ and $Y=(y_1,\ldots,y_{L_Y})$, the local cost is $c_{ij}=\|x_i-y_j\|_2$. We compute
\begin{align}
D_{ij}&=c_{ij}+\min\{D_{i-1,j},D_{i,j-1},D_{i-1,j-1}\},\label{eq:dtw}\\
d(X,Y)&=\frac{D_{L_X,L_Y}}{L_X+L_Y},\label{eq:norm}
\end{align}
with $D_{00}=0$ and the remaining boundary entries infinite. The denominator is the sum of sequence lengths, not the alignment-path length. The statistic is in meters and measures position discrepancy under unconstrained temporal warping; it omits orientation, contact, and force, so it is a geometric distance, not a tracking error.

Among the 68 clean configurations with two successful and two failed policies, Fig.~\ref{fig:overview} shows the one whose three stratum medians best match the pooled L0 medians; population results use all eligible pairs, not this example. Our primary estimate is the empirical distribution of $d$ within each outcome stratum and level. An SS median therefore describes only pairs that both succeeded, not every pair on the grid; strata also differ in task and policy-pair composition. We retain those counts and inspect task-, template-, and policy-pair-specific results alongside pooled summaries.

\subsection{Successful Controls and State-Variation Baseline}\label{sec:design-demo}
For each clean SS pair, keep the first policy in alphabetical archive-key order as an anchor and replace its partner with a successful trajectory of the same partner policy and template. The within-task control selects another initial state and the across-task control another task, both retaining L0 and successful outcomes. One eligible partner per control, sampled with seed 20260908, yields 2,588 matched triples, and twenty fixed seeds assess partner-choice sensitivity. These controls separate exact state matching from the joint effect of task geometry and goal specification.

A second control asks whether cross-policy discrepancy exceeds within-policy state variation. For anchor policy $A$ at state $s$, select another state $s'$ at which \emph{both} $A$ and partner policy $B$ succeed on the same task and template. Compare
\begin{align}
d_{\mathrm{within}}&=d(X_{A,s},X_{A,s'}),\qquad
 d_{\mathrm{cross}}=d(X_{A,s},X_{B,s'}),\nonumber\\
\delta&=d_{\mathrm{cross}}-d_{\mathrm{within}}.\label{eq:statecontrol}
\end{align}
The same replacement state is used on both sides. Four SS pairs have no eligible common state, leaving 2,584 comparisons; the primary seed is 20260909 and twenty seeds test partner-sampling sensitivity. The fixed alphabetical anchor makes this control directional, measuring state variation among archived successes. Swapping which policy is retained leaves the pooled increment positive.

A third reference set is external: the official LIBERO demonstrations for these ten tasks, fifty per task and 500 in total, from a pinned dataset revision checked file by file against published hashes. The stored demonstrations use different upstream trimming from our archive, so this comparison resamples both sides to fifty equally spaced normalized-time positions before applying the same recurrence. Its values are consequently not term-by-term comparable with the original-sampling distances above and are never subtracted from them. For each execution we take the median distance over the reference set, then report the median over executions, using same-task and other-task references and the leave-one-out distance among demonstrations as the reference scale. No state matching to demonstrations is available, and we make no claim about which files trained any checkpoint.

\subsection{Dependence, Alignment, and the Observation Window}
An episode appears in multiple policy pairs, so we resample task--state blocks, not individual pair distances. Each of 2,000 bootstrap replicates draws twenty states with replacement within each of the ten fixed tasks, yielding percentile 95\% intervals conditional on those tasks. Successful controls reuse replacement trajectories across blocks, so we report descriptive medians and sampling ranges, not independent-pair confidence intervals.

To test sampling and spatial alignment, we first remove initialization and recompute distances. We then linearly resample each path at fifty equally spaced cumulative arc-length positions (same recurrence, denominator 100), reducing the influence of differing record counts and dwell times. Subtracting each resampled path's centroid additionally removes translation while preserving scale and orientation. A stricter test resamples to fifty equally spaced normalized-time positions and then restricts the warping path to a Sakoe--Chiba band; because a band on a fifty-point grid has an integer half-width, we report the realized fraction rather than the requested one. None of these operations removes task phases or endpoint constraints.

A separate diagnostic fixes the observation window at 72 control records, or 30\% of the control cap. We retain only pairs whose two episodes remain active beyond that point. After dropping ten startup records, every-fifth sampling gives fifteen positions per path; recomputing full-episode distances on the same cohort separates cohort selection from truncation. The cutoff was fixed before this reanalysis and outcomes still refer to eventual success, making this an explanatory comparison, not an online classifier.

\section{Results: Success-Associated Trajectory Proximity}
\begin{figure*}[t]
\centering\includegraphics[width=\textwidth]{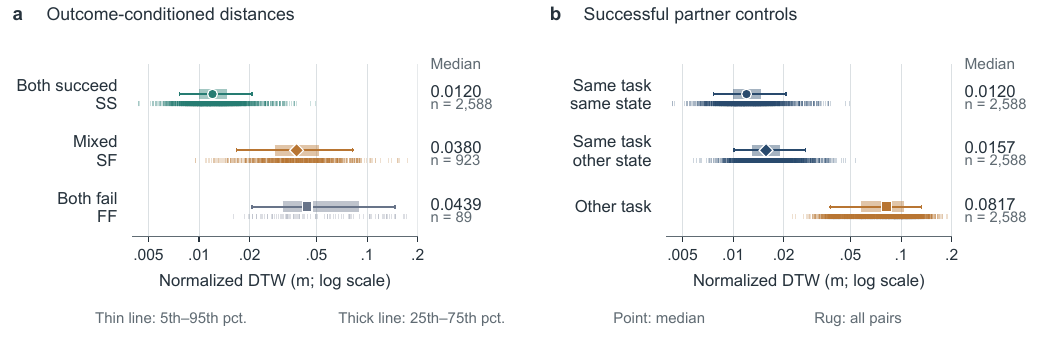}
\caption{Clean L0 distances. (a) Configuration-matched pairs by outcome. (b) Successful partner controls. Marks are defined in the figure key; both panels use a logarithmic scale and describe observed distributions, not confidence bounds.}
\label{fig:geometry}
\end{figure*}
\begin{figure}[t]
\centering\includegraphics[width=\columnwidth]{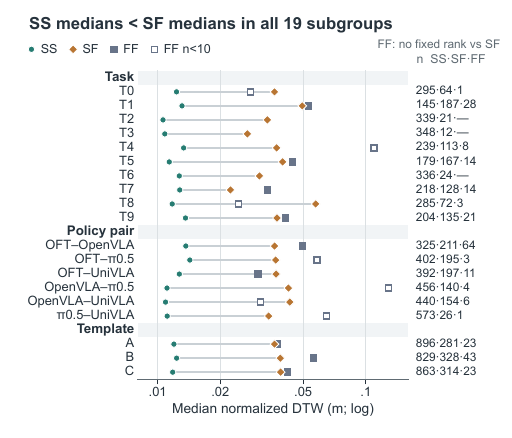}
\caption{Subgroup ordering at L0. Each row gives the SS and SF median on one logarithmic axis for ten tasks, six policy pairs and three templates from the same 3,600 pairs; SS lies left of SF in every row. FF is shown for reference (hollow: $n<10$; dash: absent) and holds no fixed rank against SF. Counts are SS$\cdot$SF$\cdot$FF; rows are subgroups, not independent samples.}
\label{fig:composite}
\end{figure}

\subsection{Successful Paths Agree More Closely}\label{sec:main}
The clean SS median is 0.0120\,m ($n=2{,}588$), compared with 0.0380\,m for SF ($n=923$) and 0.0439\,m for FF ($n=89$), as shown in Fig.~\ref{fig:geometry}(a). Task-stratified bootstrap intervals for the median contrasts are [0.0235,0.0284]\,m for SF minus SS and [0.0245,0.0506]\,m for FF minus SS.

The SS--SF ordering holds in all ten tasks, six policy pairs and three templates (Fig.~\ref{fig:composite}), and in nine representations (Section~\ref{sec:alignment}). Policy-pair SS medians range from 0.0109 to 0.0143\,m and SF medians from 0.0341 to 0.0431\,m, and the per-task counts show the pooled contrast is not a single task-invariant effect size. The SS--SF ordering also holds in each of the three policy pairs that exclude OFT (Fig.~\ref{fig:composite}). The result is a stable ordering over the representations examined, not a universal distance scale: a common observation window attenuates the gap, and successful-control comparisons reveal residual policy differences.

The distributions overlap despite their different medians: the empirical 5th--95th percentile ranges are 0.0076--0.0207\,m for SS and 0.0167--0.0820\,m for SF, so a single distance in their overlap does not identify a joint outcome: the finding concerns distributions, not a per-path classifier.

FF evidence is far less balanced: 64 of its 89 pairs are OpenVLA--OFT, other pairs contribute 1--11 each, and three tasks have no FF pair at all (Fig.~\ref{fig:composite}). Excluding that dominant pair leaves 25 FF pairs with a median of 0.0381\,m. Within tasks, the FF median already falls below the SF median at the two tasks holding one and three FF pairs, and Section~\ref{sec:alignment} shows the same reversal at the pooled level under other representations. We therefore treat FF as exploratory: it is more separated than SS, but it supports neither a stable SS$<$SF$<$FF ordering nor a taxonomy of failure mechanisms.

Conditioning matters even before comparing distances. Pooling pairs by agreeing outcome mixes the SS and FF distributions, placing weight $w=n_{\mathrm{SS}}/(n_{\mathrm{SS}}+n_{\mathrm{FF}})$ on the successful component. At L0, $w=0.967$, so a low pooled distance characterizes successful execution and cannot be read as evidence of a common failed path.

\subsection{Task Identity and Exact State Matching}\label{sec:demo}
The successful-control medians are 0.0120\,m for the original task/state match, 0.0157\,m for another state within the same task, and 0.0817\,m for another task (Fig.~\ref{fig:geometry}(b)). Across twenty partner samplings the within-task median lies between 0.0155 and 0.0158\,m and the across-task median between 0.0798 and 0.0835\,m, so the larger between-task contrast is stable under these partner choices.

The interpretation is joint: replacing a task changes the arrangement, required motion, and instruction goal together, so shared demonstrations, learned priors, and manipulation constraints may all contribute. Within this archive, successful execution is more geometrically consistent within a task than across tasks.

The official demonstrations give an external reference at matched task identity. Under normalized-time resampling, successful executions ($n=2{,}033$) have a median same-task reference distance of 0.0167\,m, against 0.0164\,m for leave-one-out distances among the 500 demonstrations; other-task values are 0.0837 and 0.0841\,m respectively, and failed executions ($n=367$) sit at 0.0374\,m from same-task demonstrations. These similar magnitudes are compatible with task-associated positional geometry, not statistical equivalence. The comparison does not separate training-data overlap, task constraints, and termination rules, and it establishes no checkpoint's training-set membership. The demonstration comparison provides a reference scale within the resampled metric, rather than a tolerance for policy replacement.

\subsection{Residual Policy Differences at a Common Replacement State}
The common-state control in Eq.~\ref{eq:statecontrol} gives same-policy and cross-policy medians of 0.0133 and 0.0156\,m over 2,584 comparisons. The median paired difference is 0.0021\,m (distinct from the 0.0023\,m gap between the two medians); twenty partner samplings span 0.00195--0.00219\,m.

\begin{table}[t]
\centering
\caption{Successful state-variation controls. Within and cross denote the two distances in Eq.~\ref{eq:statecontrol}; $\widetilde\delta$ is the median of paired differences, not a difference of medians. Distances are in meters. The first listed policy is the anchor.}
\label{tab:statevariation}
\setlength{\tabcolsep}{2pt}
\begin{tabular*}{\columnwidth}{@{\extracolsep{\fill}}lS[table-format=4.0]*{3}{S[table-format=1.4]}@{}}
\toprule
Anchor / partner & {$n$} & {Within} & {Cross} & {$\widetilde\delta$} \\
\midrule
OFT / OpenVLA & 323 & 0.0117 & 0.0169 & 0.0051 \\
OFT / $\pi_{0.5}$ & 401 & 0.0124 & 0.0176 & 0.0059 \\
OFT / UniVLA & 391 & 0.0121 & 0.0162 & 0.0043 \\
OpenVLA / $\pi_{0.5}$ & 456 & 0.0142 & 0.0147 & 0.0007 \\
OpenVLA / UniVLA & 440 & 0.0144 & 0.0146 & 0.0003 \\
$\pi_{0.5}$ / UniVLA & 573 & 0.0141 & 0.0146 & 0.0006 \\
Pooled & 2584 & 0.0133 & 0.0156 & 0.0021 \\
\bottomrule
\end{tabular*}
\end{table}

The pooled increment hides policy-pair structure (Table~\ref{tab:statevariation}): the three OFT-anchored comparisons have larger increments than the other three. The OFT--OpenVLA increment (0.0051\,m) arises between two deterministic decoders, so it cannot be attributed to inference-sampling variance, while the smallest increments involve a stochastic policy. A pre-registered repeat experiment tests this directly: 1{,}200 executions rerun every policy five times at each of ten tasks, three initial states and two templates under five fixed seed labels. Within a configuration the non-sampling decoders reproduce byte-identical paths (within-policy median 0 over 480 and 380 successful rerun pairs), while the stochastic policies do not (0.0053\,m over 596 pairs for UniVLA, 0.0067\,m over 588 for $\pi_{0.5}$). Across the 265 of 360 cells where both policies succeeded at least twice, the cross-policy median exceeds the mean of the two within-policy medians by 0.0082\,m (state-block 95\% interval [0.0077,0.0085]\,m). Cross-policy separation thus exceeds rerun variability, although the result remains conditional on these tasks, states, seed labels, and success-dependent eligibility. Because the anchor convention is directional these are not a symmetric ranking, but they establish that shared successful geometry coexists with measurable, heterogeneous policy differences: low cross-policy distance does not imply equivalence.

\subsection{Sensitivity to Sampling and Spatial Alignment}\label{sec:alignment}
Successful clean executions use a median of 102 control actions while failures all use 240, so full-episode comparisons span different durations. Removing initialization, resampling at equal arc length, and additionally centering each path all leave SS below SF (Table~\ref{tab:sensitivity}).

\begin{table}[t]
\centering
\caption{Clean metric sensitivity: median normalized DTW (m). SS/SF/FF counts are 2,588/923/89. State/task columns replace the successful partner within/across tasks, using the same 2,588 triples.}
\label{tab:sensitivity}
\setlength{\tabcolsep}{2pt}
\begin{tabular*}{\columnwidth}{@{\extracolsep{\fill}}l*{5}{S[table-format=1.4]}@{}}
\toprule
Metric & {SS} & {SF} & {FF} & {State} & {Task} \\
\midrule
Original & 0.0120 & 0.0380 & 0.0439 & 0.0157 & 0.0817 \\
No startup & 0.0132 & 0.0401 & 0.0458 & 0.0163 & 0.0885 \\
Arc-length & 0.0117 & 0.0301 & 0.0478 & 0.0146 & 0.0794 \\
+ Centering & 0.0094 & 0.0305 & 0.0375 & 0.0109 & 0.0556 \\
\bottomrule
\end{tabular*}
\end{table}

Centering reduces the across-task successful-control median from 0.0794 to 0.0556\,m, still above the same-task/state median of 0.0094\,m, so translation contributes to but does not account for the contrast.

Restricting temporal warping tests whether the contrast is an artifact of free alignment. On the fifty-point normalized-time grid, integer half-widths of 9, 4 and 2 realize band fractions of 18.37\%, 8.16\% and 4.08\%; we quote realized fractions because a band set as a percentage of sequence length can be widened by a length difference until it no longer binds. We compare nine representations: unbanded length-sum normalization (dividing by the summed sample counts), unbanded alignment-path averaging (dividing by the number of aligned cells), synchronous same-index comparison (mean pointwise distance with no warping at all), and the first two under each of the three bands. Across all nine the SS median stays within 0.0115--0.0378\,m and SF within 0.0368--0.1696\,m, and SS lies below SF in every task and policy pair. The SF/SS ratio nevertheless ranges from 2.53 to 6.13, since narrowing the band raises the SF median sharply once long temporal detours are forbidden. The direction is therefore robust to these choices while the magnitude is a property of the representation, and no single multiple should be read as an effect size. FF exceeds SF in the two unbanded resampled normalizations but falls below it under synchronous comparison and all six banded settings, which is why we report no three-stratum ordering.

\subsection{A Common Control Window Attenuates the Gap}
The 72-control diagnostic retains 3,503 pairs (2,496 SS, 918 SF, all 89 FF), excluding 97 whose episodes have already ended. Full-episode medians on this cohort are almost unchanged, while the SF--SS median contrast contracts markedly (Table~\ref{tab:prefix}). The fixed-prefix SF-minus-SS contrast is 0.01007\,m (task-stratified 95\% interval [0.00842,0.01172]\,m); FF-minus-SS is 0.02005\,m ([0.01494,0.02875]\,m).

\begin{table}[t]
\centering
\caption{Observation-window comparison on the same active cohort. Entries are median DTW (m). Full retains initialization; trimmed removes it; prefix uses the first 72 control records. Both episodes are active beyond the cutoff.}
\label{tab:prefix}
\setlength{\tabcolsep}{2pt}
\begin{tabular*}{\columnwidth}{@{\extracolsep{\fill}}lS[table-format=4.0]*{3}{S[table-format=1.4]}@{}}
\toprule
Outcome & {$n$} & {Full} & {Trimmed} & {Prefix} \\
\midrule
SS & 2496 & 0.0120 & 0.0132 & 0.0140 \\
SF & 918 & 0.0380 & 0.0402 & 0.0240 \\
FF & 89 & 0.0439 & 0.0458 & 0.0340 \\
\bottomrule
\end{tabular*}
\end{table}

The SS--SF ordering remains in all ten tasks and six policy pairs, but the pooled contrast is only about 39\% of its same-cohort full-episode value. Matching observed prefix length while conditioning on survival to 72 records preserves the ordering, so unequal duration and later behavior do not create the entire separation, though full-episode geometry magnifies it. This is an explanatory comparison on a survived-to-72 cohort, not a failure-onset estimate.

\subsection{Residual Outcome Association After Endpoint and Duration Adjustment}
Success shortens episodes and pulls endpoints together, so the raw ordering could partly reflect duration and endpoint proximity. These are consequences of outcome, not pre-treatment confounders, and endpoint mismatch shares terms with $d$, so this is a mechanical adjustment, not causal control. Regressing $d$ on SF/FF indicators over the 3{,}600 clean pairs and adding regressors in stages, the SF coefficient falls from $0.0294$ (raw) to $0.0277$ (task and pair fixed effects), $0.0156$ (adding endpoint mismatch), and $0.0048$ (adding length gap and mean length); FF falls from $0.0508$ to $0.0141$. Both stay positive with task--state cluster-bootstrap 95\% intervals excluding zero at every stage (full model SF $[0.0017,0.0077]$, FF $[0.0034,0.0260]$).

The full specification is strongly collinear, which bounds what it can settle: measured over the complete design rather than the continuous covariates alone, variance inflation factors reach $17.1$ for mean length, $11.9$ for the length gap and $10.1$ for the SF indicator, because outcome, duration, and endpoint proximity carry largely overlapping information. The surviving coefficient is accordingly specification-dependent rather than an identified effect: two pre-specified simpler duration controls (length gap only, mean length only) leave SF at $0.0082$ and $0.0058$, both above the full-model value. A permutation test reassigning outcomes within each task--state--template block (preserving per-block success counts and pair structure) places the observed SF$-$SS median contrast ($0.0260$) beyond all 5{,}000 permutations (null median $0.0130$; $p=0.0002$; per-task Cliff's $\delta$ $0.76$--$0.997$). The association therefore does not vanish under the measured endpoint and duration variables, and Section~\ref{sec:alignment} shows it is not an artifact of unconstrained warping: these data establish a residual positional association, not a decomposition of it.

\section{Performance and Composition under Visual Stress}
\subsection{The Clean Ranking Does Not Persist}
The clean Spatial ordering is $\pi_{0.5}>\mathrm{UniVLA}>\mathrm{OpenVLA}>\mathrm{OFT}$. At L3 it becomes $\pi_{0.5}>\mathrm{OFT}>\mathrm{UniVLA}>\mathrm{OpenVLA}$ (Table~\ref{tab:success}). UniVLA falls from 577/600 to 66/600 successes and OFT from 404/600 to 192/600, and the surviving L3 successes concentrate on a few tasks (e.g., 31 of OpenVLA's 41 occur on T8).

\begin{table}[t]
\centering
\caption{Empirical success rates, 600 rollouts per policy and level. Object is evaluated only for $\pi_{0.5}$.}
\label{tab:success}
\setlength{\tabcolsep}{2pt}
\begin{tabular*}{\columnwidth}{@{\extracolsep{\fill}}l*{5}{S[table-format=1.3]}@{}}
\toprule
Policy / suite & {L0} & {L1} & {L2} & {L3} & {L4} \\
\midrule
OpenVLA / Spatial & 0.762 & 0.837 & 0.683 & 0.068 & 0.000 \\
OFT / Spatial & 0.673 & 0.650 & 0.522 & 0.320 & 0.013 \\
UniVLA / Spatial & 0.962 & 0.957 & 0.697 & 0.110 & 0.000 \\
$\pi_{0.5}$ / Spatial & 0.992 & 0.983 & 0.978 & 0.567 & 0.007 \\
$\pi_{0.5}$ / Object & 0.977 & 0.987 & 0.968 & 0.637 & 0.003 \\
\bottomrule
\end{tabular*}
\end{table}

This rank change applies only to the tested pipelines under the nominal operators. All four policies lose success between L0 and L3, and $\pi_{0.5}$ leads at every level up to L3. The result does not establish statistical independence between capability and robustness, which theory suggests are coupled~\cite{capability}, and the differing views, preprocessing, corruption draws, and query schedules prevent attribution to an action representation alone. L1 is not uniformly harmful---OpenVLA rises from 0.762 to 0.837---so the response is not monotonic, while at L4 all Spatial policies fall below 0.014. On Object, $\pi_{0.5}$ likewise holds through L2 and degrades at L3--L4, extending one policy's qualitative profile without replicating the four-policy analysis across suites. OFT's clean-grid success rate is below its published LIBERO-Spatial result~\cite{openvla_oft}; the evaluations use different instruction/state grids and control horizons, and this archive does not identify the source of the discrepancy, so the reported comparisons concern the recorded pipelines, not a reproduction of published leaderboard performance.

\subsection{Changing Outcomes Change the Comparison Population}
SS distance changes modestly through L2, then rises to 0.0206\,m at L3 (Table~\ref{tab:levels}). The population also turns over: an agreeing-outcome distribution changes through both its component distributions and their weights, and the 223 surviving SS pairs at L3 are not a longitudinal panel of all clean successes. Matching pair identities across levels separates composition from geometry: 195 of those 223 pairs are also SS at L0 and 28 are new, and on that shared support the median paired increase is 0.0076\,m at L3, against 0.0016\,m at L2 and 0.0003\,m at L1.

\begin{table}[t]
\centering
\caption{Normalized DTW medians (m) and counts. Each level contains 3,600 configuration-matched pairs.}
\label{tab:levels}
\setlength{\tabcolsep}{2pt}
\begin{tabular*}{\columnwidth}{@{\extracolsep{\fill}}l*{3}{S[table-format=1.4]S[table-format=4.0]}@{}}
\toprule
& \multicolumn{2}{c}{SS} & \multicolumn{2}{c}{SF} & \multicolumn{2}{c}{FF} \\
\cmidrule(lr){2-3}\cmidrule(lr){4-5}\cmidrule(lr){6-7}
Level & {$d$} & {$n$} & {$d$} & {$n$} & {$d$} & {$n$} \\
\midrule
L0 & 0.0120 & 2588 & 0.0380 & 923 & 0.0439 & 89 \\
L1 & 0.0122 & 2652 & 0.0372 & 864 & 0.0443 & 84 \\
L2 & 0.0133 & 1928 & 0.0354 & 1328 & 0.0301 & 344 \\
L3 & 0.0206 & 223 & 0.0382 & 1471 & 0.0446 & 1906 \\
L4 & \multicolumn{1}{c}{--} & 0 & 0.0580 & 36 & 0.0678 & 3564 \\
\bottomrule
\end{tabular*}
\end{table}

SS is the least separated group wherever it exists, but SF and FF do not follow one universal hierarchy: at L2 FF has a lower median than SF, and at L4 there are no SS pairs and 3,564 of the 3,600 comparisons are FF, leaving 36 SF pairs. Distances and sample composition must therefore be read together.

\subsection{Instruction Templates: a Limited Clean Contrast}
At L0, template A exceeds B by 0.160 for OpenVLA across 200 matched task--state cells (exact McNemar $p=4.22\times10^{-5}$, Holm-adjusted $1.69\times10^{-4}$); OFT, UniVLA, and $\pi_{0.5}$ differences are $-0.035$, $+0.005$, $+0.015$ (adjusted $p$ $0.75$, $1.00$, $0.75$). One execution per cell cannot separate wording from inference variability, so this is exploratory support for systematic language evaluation~\cite{liberopara,multilingual}.

\section{Discussion and Research Implications}
\subsection{What the Geometry Establishes}
The measurements establish a specific regularity whose direction, but not size, is stable across the representations examined. Within successful executions the structure is primarily task-associated, while changing task also changes goal geometry and instruction; it is not policy equivalence, since a within-policy baseline exposes a small, heterogeneous cross-policy increment. Plausible sources include a shared bowl-placement form, common reach/grasp/place constraints, and overlapping demonstrations. The reference in Section~\ref{sec:demo} places successful executions at the scale of the demonstrations' own mutual spread, consistent with all three and separating none. Path agreement is also not execution quality: similar positions can accompany different orientations, contacts, forces, or unnecessary motion, absent from our position-only distance~\cite{uncertaintyquality}, nor correct language grounding~\cite{liberocf}.

\subsection{Implications for Cross-Policy Evaluation}
For offline cross-policy comparison, success rates and outcome-conditioned path distances answer different questions: a low SS distance does not establish interchangeability, and cross-level changes must be read together with changes in which pairs enter each outcome stratum. Cross-policy trajectory reports should state the outcome-conditioned population, show support by task and policy pair, preserve outcomes in partner controls, separate complete episodes from common windows, and report the alignment constraint with the resulting magnitude. These retrospective analyses are not deployable monitors: prospective use requires a decision clock, task-held-out evaluation, and threshold calibration~\cite{safe,hideseek}.

\subsection{Scope and Reproducibility}
The evidence remains conditional on four checkpoints, ten simulated tasks, and twenty initial states per task, with directional anchoring and trajectory reuse in the controls. The main archive provides one execution per cell, so repeated-inference behaviour rests on the separate repeat experiment, whose eligibility is success-dependent. The demonstration reference compares magnitudes in a resampled representation without state matching. Realized corruption seeds and queried images are absent. Both-failure comparisons are the thinnest evidence here and should not be extended beyond the recorded pairs.

\section{Conclusion}
Across four VLA policy pipelines on LIBERO-Spatial, executions that both succeed are more similar in end-effector positional geometry than executions with mixed outcomes. That direction persists under band-limited alignment, a common observation window, and endpoint and duration adjustment, while its magnitude depends on the representation. Within successful executions, matched task and state replacements locate the agreement as largely task-associated, and a within-policy baseline still reveals residual policy differences; both-failure comparisons stay too thinly supported to order reliably. Cross-policy trajectory evaluation is thus most informative when conditioning, control matching, observation duration, alignment constraints, and sample support are reported together. To support reproduction and community use, our code and data will be made publicly available upon publication.

\section*{Acknowledgment}
OpenAI Codex and GPT-5.6-sol assisted with drafting, code for analysis and figures, and language refinement; GLM assisted earlier drafting; MiniMax speech-2.8-hd generated the video narration. Rollouts were collected by the evaluated policies in simulation; all scientific content, analysis, and conclusions are the authors' own.
\bibliographystyle{IEEEtran}
\bibliography{references,bibstyle_control}

@IEEEtranBSTCTL{paper:bstcontrol,
  CTLuse_forced_etal = "yes",
  CTLmax_names_forced_etal = "6",
  CTLnames_show_etal = "1"
}

@article{benchmarkaudit,
 author = {Tianchong Jiang and Xiangshan Tan and Samuel Wheeler and Luzhe Sun and Tewodros W. Ayalew and Matthew Walter},
 doi = {10.48550/arXiv.2606.04233},
 eprint = {2606.04233},
 journal = {arXiv preprint arXiv:2606.04233},
 title = {What Are We Actually Benchmarking in Robot Manipulation?},
 year = {2026}
}

@article{capability,
 author = {Tai, Jianwei},
 doi = {10.48550/arXiv.2605.25889},
 eprint = {2605.25889},
 journal = {arXiv preprint arXiv:2605.25889},
 title = {Capability and Robustness Cannot Both Be Free: An Information-Theoretic Bound for Vision-Language-Action Models},
 year = {2026}
}

@article{dtw,
 author = {Sakoe, Hiroaki and Chiba, Seibi},
 doi = {10.1109/TASSP.1978.1163055},
 journal = {IEEE Transactions on Acoustics, Speech, and Signal Processing},
 number = {1},
 pages = {43--49},
 title = {Dynamic Programming Algorithm Optimization for Spoken Word Recognition},
 volume = {26},
 year = {1978}
}

@article{experiences,
 author = {Zhang, Yihao and Qi, Yuankai and Zheng, Xi},
 doi = {10.48550/arXiv.2511.11298},
 eprint = {2511.11298},
 journal = {arXiv preprint arXiv:2511.11298},
 title = {Experiences from Benchmarking Vision-Language-Action Models for Robotic Manipulation},
 year = {2025}
}

@article{featurefollowing,
 author = {Kr{\"u}ger, Martin and Salem, Mahmoud and Reischl, Markus},
 doi = {10.1007/s11063-026-11860-3},
 journal = {Neural Processing Letters},
 pages = {41},
 title = {Assessment of a Fine-Tuned Vision-Language-Action Model for Robotic Feature-Following Inspection},
 volume = {58},
 year = {2026}
}

@article{hideseek,
 author = {Park, Seongheon and Li, Wendi and Oh, Changdae and Yeh, Samuel and Kira, Zsolt and Hagenow, Michael and Li, Sharon},
 doi = {10.48550/arXiv.2605.30834},
 eprint = {2605.30834},
 journal = {arXiv preprint arXiv:2605.30834},
 title = {Hide-and-Seek in Trajectories: Discovering Failure Signals for {VLA} Runtime Monitoring},
 year = {2026}
}

@inproceedings{libero,
 author = {Liu, Bo and Zhu, Yifeng and Gao, Chongkai and Feng, Yihao and Liu, Qiang and Zhu, Yuke and Stone, Peter},
 booktitle = {Advances in Neural Information Processing Systems},
 doi = {10.52202/075280-1939},
 eprint = {2306.03310},
 pages = {44776--44791},
 title = {{LIBERO}: Benchmarking Knowledge Transfer for Lifelong Robot Learning},
 volume = {36},
 year = {2023}
}

@article{liberocf,
 author = {Yu Fang and Yuchun Feng and Dong Jing and Jiaqi Liu and Yue Yang and Zhenyu Wei and Daniel Szafir and Mingyu Ding},
 doi = {10.48550/arXiv.2602.17659},
 eprint = {2602.17659},
 journal = {arXiv preprint arXiv:2602.17659},
 title = {When Vision Overrides Language: Evaluating and Mitigating Counterfactual Failures in {VLAs}},
 year = {2026}
}

@article{liberopara,
 author = {Kim, Chanyoung and Kim, Minwoo and Kang, Minseok and Kim, Hyunwoo and Jung, Dahuin},
 doi = {10.48550/arXiv.2603.28301},
 eprint = {2603.28301},
 journal = {arXiv preprint arXiv:2603.28301},
 title = {{LIBERO-Para}: A Diagnostic Benchmark and Metrics for Paraphrase Robustness in {VLA} Models},
 year = {2026}
}

@inproceedings{liberoplus,
 author = {Fei, Senyu and Wang, Siyin and Shi, Junhao and Dai, Zihao and Cai, Jikun and Qian, Pengfang and Ji, Li and He, Xinzhe and Zhang, Shiduo and Fei, Zhaoye and Fu, Jinlan and Gong, Jingjing and Qiu, Xipeng},
 booktitle = {Proceedings of the IEEE/CVF Conference on Computer Vision and Pattern Recognition (CVPR)},
 pages = {38574--38583},
 title = {{LIBERO-Plus}: A Progressive Robustness Benchmark for Visual-Language-Action Models},
 year = {2026}
}

@inproceedings{multilingual,
 author = {Dong, Xuan  and
Han, Zhe  and
Niu, Tianhao  and
Zhu, Qingfu  and
Che, Wanxiang},
 booktitle = {Proceedings of the 64th Annual Meeting of the {A}ssociation for {C}omputational {L}inguistics (Volume 1: Long Papers)},
 doi = {10.18653/v1/2026.acl-long.2066},
 pages = {44615--44629},
 publisher = {Association for Computational Linguistics},
 title = {When Does Language Matter? Multilingual Instructions Reveal Step-wise Language Sensitivity in Vision-Language-Action Models},
 year = {2026}
}

@article{nebula,
 author = {Jierui Peng and Yanyan Zhang and Yicheng Duan and Tuo Liang and Vipin Chaudhary and Yu Yin},
 doi = {10.48550/arXiv.2510.16263},
 eprint = {2510.16263},
 journal = {arXiv preprint arXiv:2510.16263},
 title = {{NEBULA}: Do We Evaluate Vision-Language-Action Agents Correctly?},
 year = {2025}
}

@inproceedings{openvla,
 author = {Kim, Moo Jin and Pertsch, Karl and Karamcheti, Siddharth and Xiao, Ted and Balakrishna, Ashwin and Nair, Suraj and Rafailov, Rafael and Foster, Ethan P and Sanketi, Pannag R and Vuong, Quan and Kollar, Thomas and Burchfiel, Benjamin and Tedrake, Russ and Sadigh, Dorsa and Levine, Sergey and Liang, Percy and Finn, Chelsea},
 booktitle = {Proceedings of The 8th Conference on Robot Learning},
 pages = {2679--2713},
 publisher = {PMLR},
 series = {Proceedings of Machine Learning Research},
 title = {{OpenVLA}: An Open-Source Vision-Language-Action Model},
 volume = {270},
 year = {2025}
}

@inproceedings{openvla_oft,
 author = {Moo Jin Kim AND Chelsea Finn AND Percy Liang},
 booktitle = {Proceedings of Robotics: Science and Systems},
 doi = {10.15607/RSS.2025.XXI.017},
 title = {Fine-Tuning Vision-Language-Action Models: Optimizing Speed and Success},
 year = {2025}
}

@article{pi05,
 author = {{Physical Intelligence} and Black, Kevin and Brown, Noah and Darpinian, James and Dhabalia, Karan and Driess, Danny and Esmail, Adnan and Equi, Michael and Finn, Chelsea and Fusai, Niccolo and Galliker, Manuel Y. and Ghosh, Dibya and Groom, Lachy and Hausman, Karol and Ichter, Brian and Jakubczak, Szymon and Jones, Tim and Ke, Liyiming and LeBlanc, Devin and Levine, Sergey and Li-Bell, Adrian and Mothukuri, Mohith and Nair, Suraj and Pertsch, Karl and Ren, Allen Z. and Shi, Lucy Xiaoyang and Smith, Laura and Springenberg, Jost Tobias and Stachowicz, Kyle and Tanner, James and Vuong, Quan and Walke, Homer and Walling, Anna and Wang, Haohuan and Yu, Lili and Zhilinsky, Ury},
 doi = {10.48550/arXiv.2504.16054},
 eprint = {2504.16054},
 journal = {arXiv preprint arXiv:2504.16054},
 title = {{$\pi_{0.5}$}: a Vision-Language-Action Model with Open-World Generalization},
 year = {2025}
}

@inproceedings{safe,
 author = {Gu, Qiao and Ju, Yuanliang and Sun, Shengxiang and Gilitschenski, Igor and Nishimura, Haruki and Itkina, Masha and Shkurti, Florian},
 booktitle = {Advances in Neural Information Processing Systems},
 doi = {10.52202/085713-1337},
 eprint = {2506.09937},
 pages = {40041--40076},
 title = {{SAFE}: Multitask Failure Detection for Vision-Language-Action Models},
 volume = {38},
 year = {2025}
}

@article{triinfo,
 author = {Yang, Jinghan and Zhang, Yunchao and {Wang Yuan} and {Haolun Wan} and Zhang, Jiaming and Hu, Zhengyang and Yang, Yanchao},
 doi = {10.48550/arXiv.2606.19998},
 eprint = {2606.19998},
 journal = {arXiv preprint arXiv:2606.19998},
 title = {{Tri-Info}: Generalizable, Interpretable Failure Prediction for {VLA} Models via Information Theory},
 year = {2026}
}

@article{uncertaintyquality,
 author = {Pablo Valle and Chengjie Lu and Shaukat Ali and Aitor Arrieta},
 doi = {10.48550/arXiv.2507.17049},
 eprint = {2507.17049},
 journal = {arXiv preprint arXiv:2507.17049},
 title = {Evaluating Uncertainty and Quality of Vision-Language-Action-enabled Robots},
 year = {2025}
}

@inproceedings{univla,
 author = {Qingwen Bu AND Yanting Yang AND Jisong Cai AND Shenyuan Gao AND Guanghui Ren AND Maoqing Yao AND Ping Luo AND Hongyang Li},
 booktitle = {Proceedings of Robotics: Science and Systems},
 doi = {10.15607/RSS.2025.XXI.014},
 title = {Learning to Act Anywhere with Task-centric Latent Actions},
 year = {2025}
}

@article{vlatest,
 author = {Wang, Zhijie and Zhou, Zhehua and Song, Jiayang and Huang, Yuheng and Shu, Zhan and Ma, Lei},
 doi = {10.1145/3729343},
 journal = {Proceedings of the ACM on Software Engineering},
 pages = {1615--1638},
 title = {{VLATest}: Testing and Evaluating Vision-Language-Action Models for Robotic Manipulation},
 volume = {2},
 year = {2025}
}

@article{wording,
 author = {Woo, Jihwan},
 doi = {10.22541/au.177497139.90709706/v1},
 journal = {Authorea preprint},
 title = {Task-Dependent Sensitivity of {VLA} Models to Instruction Wording},
 year = {2026}
}

@article{roboeval,
 author = {Wang, Yi Ru and Ung, Carter and Tan, Christopher and Tannert, Grant and Duan, Jiafei and Li, Josephine and Le, Anh and Oswal, Rishabh and Grotz, Markus and Pumacay, Wilbert and Deng, Yuquan and Krishna, Ranjay and Fox, Dieter and Srinivasa, Siddhartha},
 doi = {10.48550/arXiv.2507.00435},
 eprint = {2507.00435},
 journal = {arXiv preprint arXiv:2507.00435},
 title = {{RoboEval}: Where Robotic Manipulation Meets Structured and Scalable Evaluation},
 year = {2026}
}

@article{distractedrobot,
 author = {Rasouli, Amir and Alban, Montgomery and Pakdamansavoji, Sajjad and Li, Zhiyuan and Zhang, Zhanguang and Wu, Aaron and Zhao, Xuan},
 doi = {10.48550/arXiv.2511.22780},
 eprint = {2511.22780},
 journal = {arXiv preprint arXiv:2511.22780},
 title = {{Distracted Robot}: How Visual Clutter Undermine Robotic Manipulation},
 year = {2025}
}
\end{document}